\documentclass[pdflatex,sn-mathphys]{sn-jnl}

\jyear{2021}%
\usepackage{booktabs}
\usepackage{multirow}
\usepackage{longtable}
\usepackage{amsmath}
\usepackage{graphicx}
\usepackage{longtable}
\theoremstyle{thmstyleone}%
\theoremstyle{thmstyletwo}%

\theoremstyle{thmstylethree}%

\begin{document}

\title[Reverse Survival Model (RSM)]{Reverse Survival Model (RSM): A Pipeline for Explaining Predictions of Deep Survival Models}


\author[1,2]{\fnm{Mohammad R.} \sur{Rezaei}}\email{mohammadreza.rezaei@orthobiomed.ca}

\author[1,3]{\fnm{Reza} \sur{Saadati Fard}}\email{saadatifard.reza@gmail.com}

\author[1]{\fnm{Ebrahim} \sur{Pourjafari}}\email{ebrahim.pourjafari@orthobiomed.ca}

\author[1]{\fnm{Navid} \sur{Ziaei}}\email{navid.ziaei@orthobiomed.ca}
\author[1]{\fnm{Amir} \sur{Sameizadeh}}\email{amir.samiezadeh@orthobiomed.ca}
\author[1,4]{\fnm{Mohammad} \sur{Shafiee}}\email{Mohammad.Shafiee@uhn.ca}
\author[1,4]{\fnm{Mohammad} \sur{Alavinia}}\email{mohammad.alavinia@uhn.ca}
\author[1]{\fnm{Mansour} \sur{Abolghasemian}}\email{mansour.abolghasemian@orthobiomed.ca}
\author*[1]{\fnm{Nick} \sur{Sajadi}}\email{nick.sajadi@orthobiomed.ca}

\affil*[1]{ \orgname{Ortho Biomed Inc.}, \orgaddress{ \city{Toronto}, \state{ON}, \country{Canada}}}

\affil[2]{ \orgname{University of Toronto}, \orgaddress{\city{Toronto}, \state{ON}, \country{Canada}}}

\affil[3]{ \orgname{Worcester Polytechnic Institute (WPI)}, \orgaddress{\city{Worcester}, \state{MA}, \country{USA}}}

\affil[4]{ \orgname{University Health Network}, \orgaddress{\city{Toronto}, \state{ON}, \country{Canada}}}


\abstract{The aim of survival analysis in healthcare is to estimate the probability of occurrence of an event, such as a patient's death in an intensive care unit (ICU). Recent developments in deep neural networks (DNNs) for survival analysis show the superiority of these models in comparison with other well-known models in survival analysis applications. Ensuring the reliability and explainability of deep survival models deployed in healthcare is a necessity. Since DNN models often behave like a black box, their predictions might not be easily trusted by clinicians, especially when predictions are contrary to a physician's opinion. A deep survival model that explains and justifies its decision-making process could potentially gain the trust of clinicians. 
In this research, we propose the reverse survival model (RSM) framework that provides detailed insights into the decision-making process of survival models. For each patient of interest, RSM can extract similar patients from a dataset and rank them based on the most relevant features that deep survival models rely on for their predictions.
RSM acts as an add-on to a deep survival model and offers three functionalities: 1) Finding the most relevant clinical measurements for the probability density functions (PDFs) of events. 2) Categorizing patients into disjoint clusters based on the similarity of their survival PDFs. 3) Ranking similar patients based on the similarity of survival outcomes and relevant clinical measurements.
The explainability of deep survival models is rarely addressed in literature. Therefore, the RSM pipeline is a unique approach to explain the predictions of deep survival models. We validated the RSM pipeline by testing it on a synthetic dataset and MIMIC-IV, a dataset of intensive care unit (ICU) clinical observations.
Our experiments showed that given a deep survival model and a patient of interest, RSM can successfully detect similar patient records from historical data and rank them based on the similarities between their survival PDFs and the most relevant patient observations.}




\maketitle

\section{Introduction}
Survival analysis is a well-defined problem in machine learning that estimates the probability of the occurrence of an event of interest through time. An example of such an event is an organ failure in a recipient after an organ transplant, or the death of a patient admitted to an intensive care unit (ICU). The emergence of deep neural networks (DNNs) and their superior performance in the field of survival analysis \cite{nagpal2021deep, lee2018deephit,  lee2019dynamic, miscouridou2018deep} over traditional Cox-based \cite{therneau2000cox} and shallow machine learning models such as logistic regression \cite{efron1988logistic} and random survival forest (RSF) \cite{ishwaran2010consistency}  motivated healthcare industry and organizations\footnote{https://impact.canada.ca/en/challenges/deep-space-healthcare-challenge} to utilize DNN models for survival analysis. 
As new advances in DNNs become increasingly common in survival analysis applications \cite{lee2019dynamic,nagpal2021deep, katzman2018deepsurv,thiagarajan2020calibrating,ozen2019sanity,hanif2018robust,chung2020deep}, ensuring their operational reliability has become crucial. DNN models usually outperform traditional survival models in estimating the probability density functions (PDFs) of events by learning complex interconnected relationships between the observations and events \cite{rezaei2022deep}. 
The decision-making process of traditional survival models such as Cox and RSF is simple and easy to interpret by a human. On the other hand, DNNs can learn complex and interconnected relationships between features and targets. However, the internal working process of DNNs looks like a black box and therefore, is not easily interpretable by a human. Consequently, predictions of DNNs, especially in healthcare settings, might not be easily explainable or trusted.
There have been a few attempts to gain the trust of healthcare professionals on DNN predictions.
\cite{ribeiro2016should} developed an algorithm named LIME to make classifier or regression models interpretable, by adding an interpreter model such as a decision tree to identify a list of important features relevant to the predictions. While the algorithm is applied to text processing and computer vision, it is not applied to survival analysis. Applying LIME to survival analysis would be significantly harder, as there is no ground truth for the survival function. The LIME algorithm requires training at least two models at the same time, which makes the training process complex. The other drawback of LIME is that the results are subjective to a specific problem and its accuracy cannot be measured statistically.\\
\cite{yousefi2017predicting} introduced a pipeline for interpreting survival analysis results using a DNN. They used Cox partial likelihood as the cost function and back-propagated the calculated risks to the first layer of the network to determine the risk factor of each input feature corresponding to a patient prognosis. The calculated risk for each input provides an interpretation factor for the whole model. Although the model provides an approach to make a DNN model interpretable, it cannot be used when the electronic health record (EHR) cohort has longitudinal measurements, missing values, censored records, or competing risks. Though the proposed pipeline provides a ranked list of the most relevant features, it cannot identify similar patients to a patient of interest based on the similarity of outcomes or EHR records as a source of trust to predictions.\\
In the field of healthcare, identifying patients with clinical measurements and outcomes similar to a case under investigation is considered a source of trust. \cite{gallego2015bringing} developed a process that can provide similar patients from the database to an individual patient based on the past clinical decisions and clinical verdicts. \cite{sun2012supervised} used a generalized Mahalanobis distance \cite{de2000mahalanobis} for deriving similar patients based on a physician's feedback. Their method is a supervised learning approach for clustering EHR patients based on key clinical indicators. This approach does not apply to survival analysis, since the output of survival analysis is a survival function with no ground truth. \cite{li2016deep} discussed a few methods for ranking features during training a DNN for genome research, where all those methods add a regularized term to the cost function of the model to rank input features. Although this technique is simple and effective to rank input features, it needs the ground truth for DNN targets to be able to classify significant features associated with each class. Unfortunately, the assumption of availability of the ground truth is not held in many problems including survival analysis, where the ground truth is unknown, including \cite{nagpal2021deep}. \\
To the best of our knowledge, there is no unified model or framework specifically designed for the interpretation of deep survival models. A comprehensive interpretability tool for a deep survival model can be used for evaluating the reliability of predictions of the model and consequently increase the chance of acceptance of that model by healthcare professionals.
In this research, we propose a framework, reverse survival model (RSM), that provides further insights into the decision-making process of deep survival models. For each prediction, RSM extracts similar clinical measurements and ranks them based on their relevance to the predicted survival PDFs of a deep survival model. For example, RSM can provide a list of similar patients in terms of their survival PDFs and the most relevant clinical observations.
It has been shown in \cite{che2016interpretable,lee2019dynamic,katzman2018deepsurv,nagpal2021deep,miscouridou2018deep} that the estimated PDFs of DNN models usually surpass the PDFs of traditional survival models in terms of accuracy and quality. However, when it comes to individual predictions, a physician might not trust a DNN model, since the range of error for an individual prediction is unknown to the physician. In this paper, we try to address this question: \emph{"when should we trust an individual survival prediction by a DNN model?"}.  
Our response to this question is to provide a source of trust; a list of \emph{similar patients} from the history of the model, with clinical measurements and outcomes that are similar to the current prediction being made.\\
To interpret the outcomes of a deep survival model, RSM executes three major steps: 1) It finds the features that are most relevant to the predictions made by deep survival models, and 2) it categorizes patients into a few distinct clusters based on the similarity between survival predictions, and, 3) it ranks similar patients based on their survival PDFs and similarities among relevant clinical measurements. RSM applies Jensen–Shannon divergence (JSD) between survival PDFs, as the measure of similarity \cite{fuglede2004jensen}. Smaller the JSD between the PDFs of two patients, more similar the outcomes for those two patients are \cite{connor2013evaluation}. The significance of clinical measurements is measured by the Kolmogorov-Smirnov statistical test  \cite{massey1951kolmogorov}.\\
The unique advantage of RSM is that it can be applied to any deep survival models that predict the time to an event based on the clinical measurements of an EHR, where the EHR can contain longitudinal measurements, missing values, and censored records. We tested RSM on a synthetic dataset and MIMIC-IV \cite{johnson2020mimic}, a well-known ICU dataset, for survival analysis.
The results prove that for each prediction, RSM can successfully identify similar records from historical data, and then rank them, based on the degree of similarity.\\
The rest of this paper is organized as follows: The pipeline of RSM is described in Section \label{reversed_section}. Section \ref{section:cohort} introduces the datasets used for evaluating the model. Experimental results are provided in Section \ref{section:experiments}. Finally, Section \ref{section:Discussion} discusses the characteristics and limitations of RSM and concludes the paper.
\section{Methods}\label{reversed_section}
Assume that $D$ consists of a set of tuples $\{(\boldsymbol{x}_i, t_i, \delta_i)\}_{i=1}^N$, where $x_i \in \mathbb{R}^M$ is a vector of $M$ clinical measurements of an individual $i$,  $t_i$ and $\delta_i$ are time and type of an event like death, respectively. 
\begin{figure}[!tb]
		\centering
		\includegraphics[width=\linewidth]{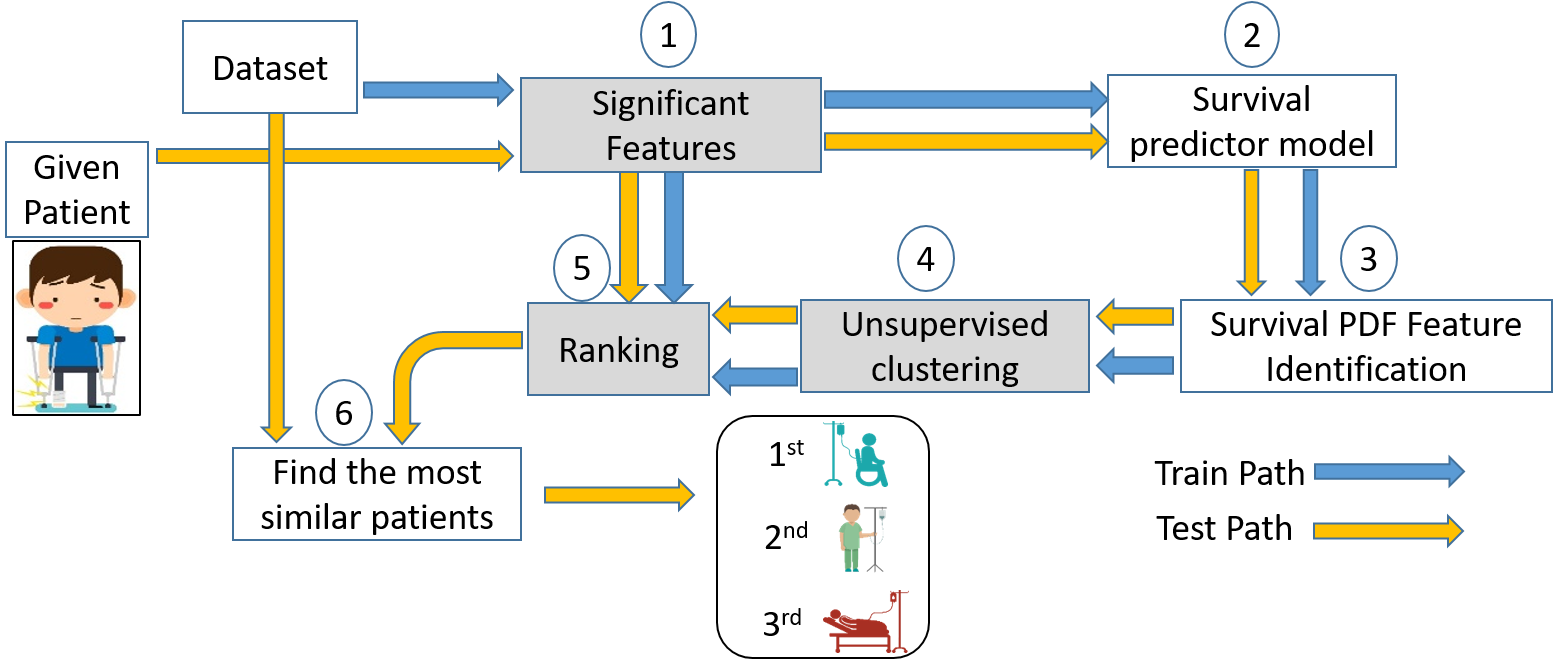}
		\caption{Schematic view of the RSM model. During the training phase, RSM learns to identify the most significant clinical measures related to the survival PDF of a patient, along with its designated cluster for similarity. For the test path, based on the similarity of survival PDFs and significant clinical measures, Unit 6 suggests the most similar patients from the dataset for a test patient.}
		\label{fig:RSM_PIPELINE}
	\end{figure}	
Given a trained deep survival model, we hypothesize that two patients with similar clinical observations would likely generate similar outcomes. Note that this is based on the assumption that clinical observations are sufficient and relevant to clinical events. As it is shown in Figure \ref{fig:RSM_PIPELINE}, RSM consists of 6 different units. Unit 1 estimates the rank of clinical measurements for the survival PDF prediction. Unit 2 consists of a deep survival model which predicts the survival PDFs. In this study, we use the Survival Seq2Seq model \cite{survival_Seq2Seq}, a state-of-the-art model for survival analysis. Unit 3 extracts statistics from the estimated PDFs. Unit 4 clusters patients with similar survival PDFs based on the outcome of unit 3. Unit 5 ranks patients in each cluster based on the similarity between their survival PDFs.\\
RSM has two phases of training and testing that utilize these six units slightly differently.
All units 1-5 are parts of the training path of RSM. In the training phase, Units 1 and 2 are trained using traditional optimization techniques for deep survival models considering a modified loss function defined in equation \ref{eq-significant-CM}. Then, unit 4 learns to cluster the patients by extracting features from the predicted survival PDFs by unit 3 (see section \ref{pdf-similarity-section} for more details). Finally, unit 6 learns to find the most significant clinical measurements and gives a similar measure that represents the closeness of the most significant clinical measurements. In the test phase, RSM takes the clinical measurements of a test patient and runs units 1 to 5 to reach a similarity score to other patients with the similar designated cluster, those with similar survival PDFs, based on the most significant clinical measurements. Finally, unit 6 ranks the most similar patients in the associated cluster to the test patient and returns them as the output.
\subsubsection*{PDFs Similarity Measure}
\label{pdf-similarity-section}
There are several measures of similarity for PDFs \cite{cha2007comprehensive}, and JSD is a well-known example of that. JSD is a symmetric measure between two PDFs \cite{fuglede2004jensen} defined by
\begin{equation}
JS_{dist}(P\parallel  Q) =\sqrt{ \frac{1}{2} D(P\parallel  Q) + \frac{1}{2} D(Q\parallel  P)},
\end{equation}
where $P$ and $Q$ are PDFs and $D(P \parallel  Q)$ is the Kullback–Leibler (KL) divergence between $P$ and $Q$ distributions.
RSM clusters patients into $K$ clusters based on their JSD similarity scores. RSM calculates the JSD distance for each pair of survival PDFs as a measure of distance. This measure is used to cluster patients. We used K-means clustering for the sake of simplicity and generalizability. However, alternative clustering algorithms can also be used to cluster patients as well. 
\subsection*{Significant Input Features (Clinical Measurements) And Their Ranking} \label{signifcant_sect}
Estimating the rank of clinical measurements with respect to a DNN prediction is crucial to the functionality of RSM. Considering all features for measuring the similarity of predictions of two patients could be computationally prohibitive. Therefore, to reduce computational complexity, RSM finds similarities among patients by only considering the features that are most relevant to survival predictions.\\
\begin{figure}[!tb]
		\centering
		\includegraphics[width=\linewidth]{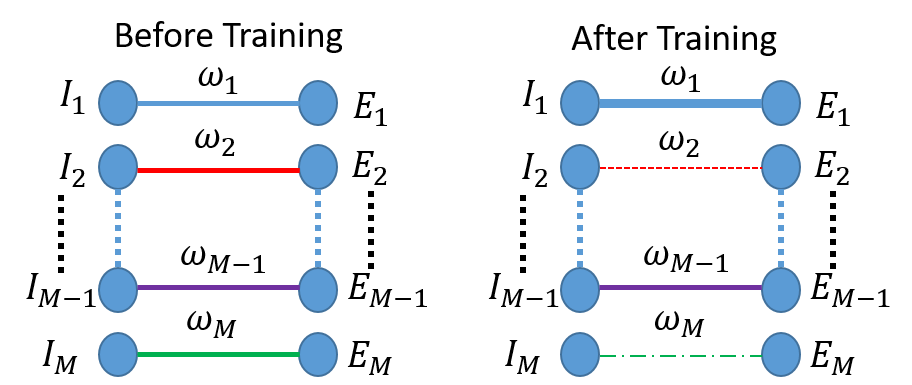}
		\caption{Estimating the significance of clinical measurements by Unit 1. A trainable weight is assigned to each clinical measurement, which represents the importance of each clinical measurement for the survival model predictions. The magnitude of each weight is proportional to the importance of the corresponding clinical measurement. RSM uses $L_1$ regularization for optimizing these weights. Consequently, insignificant measurements are assigned smaller weights, as represented in the right panel by weak connections such as $\omega_2$ and $\omega_M$ for $I_2$ and $I_M$ measurements, respectively.}
		\label{fig:fig_3}
	\end{figure}
The rank and significance of clinical measurements are estimated by assigning a trainable weight, $\omega$, to each clinical measurement, $I$, and then use the weighted ones, $E$, as the input of the survival model, $f$, as is shown in Figure \ref{fig:fig_3}.
After training, the one-to-one layer identifies the significant features by assigning a weight to each feature, $\omega_m, m={1, ...,M}$, where $M$ is the input dimension. 
The loss function of the survival model is modified to learn feature weights and is given by
\begin{equation}
\label{eq-significant-CM}
\mathcal{L'}^{(i)}=\mathcal{L}^{(i)}(y^{(i)},f_{\theta}(x^{(i)})) +\frac{\lambda}{M}\sum_{m=1}^{M} \|\omega_m\|,
\end{equation}
where the term $\mathcal{L}^{(i)}(y^{(i)},f_{\theta}(x^{(i)}))$ is the original loss function of the survival model $f_{\theta}$ for patient $x^{(i)} \in R^M$ estimating label $y^{(i)}$, and the term $\frac{\lambda}{M} \sum_{m=1}^{M} \|\omega_m\|$ is the regularization term. Here, $|.|$ represents the absolute value of weight $\omega_m$ associated with the $m^{th}$ input clinical measurement. The hyperparameter $\lambda$ is the regularization term. The $L_1$ regularization technique \cite{molchanov2017variational,chang2017dropout} is used to estimate the rank and significance of clinical measurements. The absolute value of a regularized weight represents the significance of the associated clinical measurement in the deep survival model predictions, as shown in Figure \ref{fig:fig_3}.\\
After the features are ranked, RSM uses the two-sample Kolmogorov–Smirnov (KS) statistical test \cite{massey1951kolmogorov} to identify the most and least significant clinical measurements with respect to the outcomes of the deep survival model (survival PDFs). The two-sample KS enables capturing discontinuity, heterogeneity, and dependence across data samples \cite{naaman2021tight}, which is beyond the ability of simpler statistical tests like T-test \cite{naaman2021tight}.
The Kolmogorov-Smirnov test (K-S test) compares the data with a known distribution and confirms if they have the same distribution. More specifically, the K-S test compares the distribution of the significance of all the clinical measurements, to the distribution generated by the most significant clinical measurements. We make the later distribution by selecting some features from the top-ranked features. This number incrementally increases to the point that the K-S test shows no significant difference between the distribution of significance of all clinical measurements and the most significant ones. Therefore, the identified number represents the most significant clinical measurements that represent the distribution of the all measurements. After finding the most significant clinical measurements, RSM ranks patients similar to a test patient based on the rank of the most significant clinical measurements and the cluster the predicted PDFs fall in.  
\subsubsection*{Finding The Most Similar Patients}
\label{PCA_desc_section}
Units one to five of RSM explained so far and depicted in Figure \ref{fig:RSM_PIPELINE} are optimized in the training phase of RSM. In the test phase, RSM aims to find the most similar patients from the historical dataset, based on the predicted survival PDFs of a survival analysis model. First, RSM finds the cluster of the test patient using the trained clustering model. Then, if clinical measurements are continuous, RSM  compares the values of clinical measurements against each other based on a Euclidean distance to find and rank the most similar patients. We will showcase this approach on a synthetic dataset in later sections. 
For larger datasets with hundreds of numerical and categorical features, a simple ranking approach based on Euclidean distance between the clinical measurements becomes computationally expensive. To reduce computations, RSM uses Principal component analysis (PCA) to describe the clinical measurements by a set of linearly uncorrelated principal-components \cite{abdi2010principal}. The number of the most significant principal components is relevantly smaller than the number of input clinical measurements \cite{reddy2020analysis}. Therefore, ranking becomes computationally efficient. RSM ranks patients based on the Euclidean distance of their significant feature representations in the subspace of the largest principal components.\\
Generally, the PCA analysis reduces the computational complexity of finding and ranking similar patients to test patients. Each eigenvector of PCA represents a variance measure for the selected significant clinical measurements. In other words, we apply PCA on a matrix consisting of patients in the train set, where only selected significant clinical measurements exist. For patients in each cluster resulting from K-means with J-S difference, we calculate a weighted sum using significant eigenvectors, where the weights are eigenvalues, i.e, each eigenvector is multiplied with each patient's clinical measurements and weighted with its eigenvalue. This calculation gives a score measure that represents the magnitude of the transformed patient in the significant PCA space resulting from significant clinical measurements. This score can be used to measure the similarity of patients in the PCA space. If we sort all patients in a cluster using this score and apply the same procedure to the test patient to calculate that patient's score, we can simply find the most similar patients using a binary search. For example, assume you want 10 similar patients, by finding the placement of the test patient in the cluster, choose 5 patients above and 5 patients under the found placement.\\
Despite the advantages of using PCA, there are assumptions about PCA that should be considered. PCA assumes an affine transformation among significant clinical measurements. Also, measurements should be independent and identically distributed (iid) which is a valid assumption in our analysis, i.e., we know that measurements of each patient are independent of other patients. PCA is fit optimally if these assumptions are satisfied. Otherwise, the outcome will be sub-optimal. If the affine transformation assumption between data samples does not hold, one can use kernel-PCA to take into account non-linear relationships \cite{rosipal2001kernel}. 
\section{Experiments}
\label{section:cohort}
We assessed the performance of RSM on two datasets: a synthetic dataset that partially resembles medical datasets and MIMIC-IV, a well-known ICU dataset. A detailed description of each dataset is provided in the following subsections.
\subsection{The Synthetic dataset}
\begin{figure}[!tb]
		\centering
		\includegraphics[width=.8\linewidth]{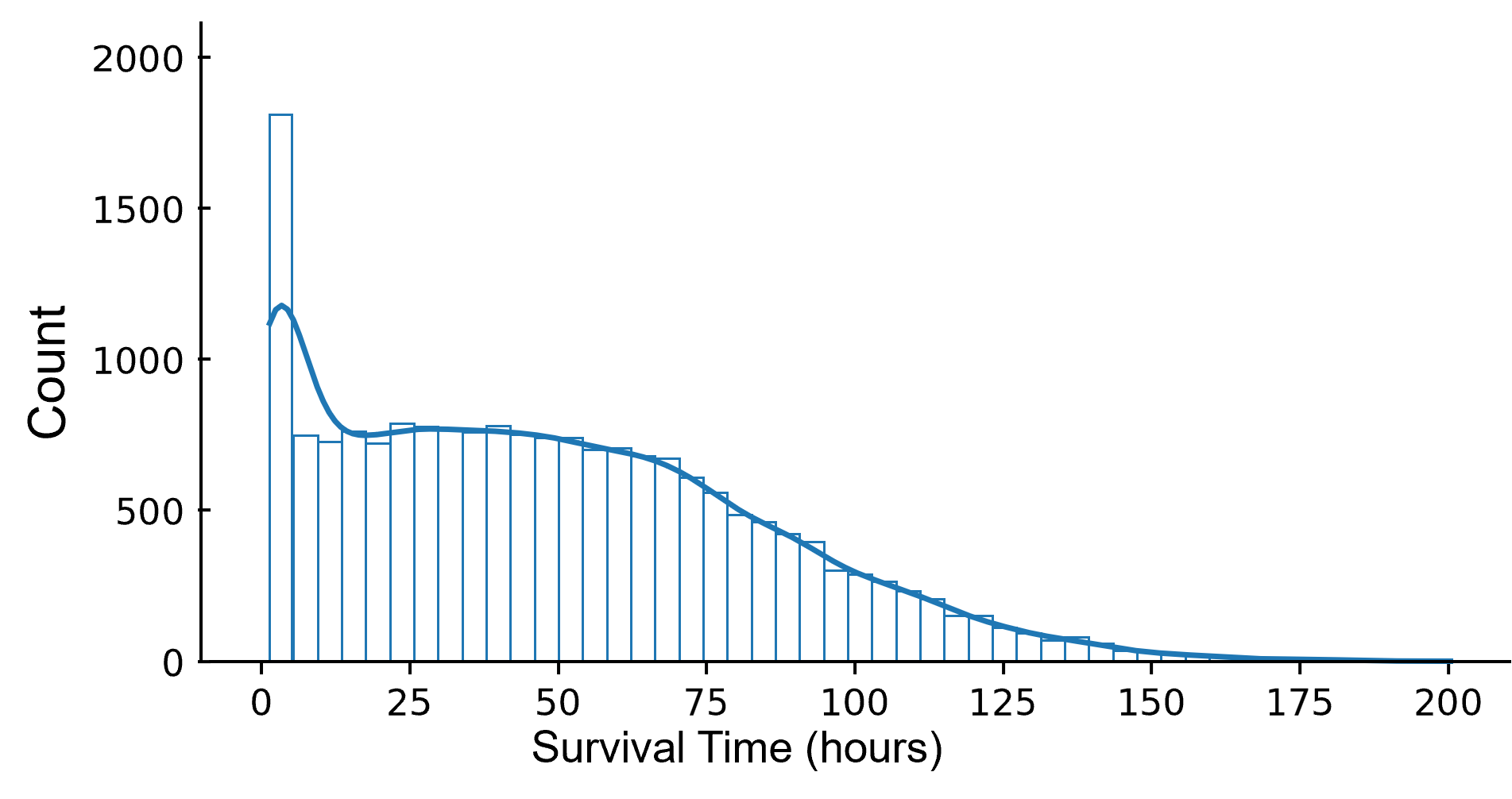}
		\caption{The histogram of survival times in the Synthetic dataset. The solid blue line shows a kernel density estimation interpolation of the bars that shows the frequency of each quantified survival time \cite{kim2012robust}.}
		\label{fig:fig_4}
	\end{figure}

To investigate the ability of RSM in interpreting the predictions of deep survival models, we created a synthetic dataset based on a statistical process. We considered $\boldsymbol{x}=(x^{1},...,x^{K})$ as a tuple of $K$ random variables, where each random variable can be considered as a clinical measurement with an independent normal distribution, $N \sim (0,\boldsymbol{I})$. We modeled the distribution of the event time, $T_i$, for each data sample $i$ as a nonlinear combination of these $K$ random variables at time index $i$ given by
\begin{equation}
\label{synthetic_data}
    T_i \sim \exp((\boldsymbol{\alpha}^T\times{(\boldsymbol{x}_{i}^{k_1})}^2+(\boldsymbol{\beta}^T\times(\boldsymbol{x}_{i}^{k_2})),
\end{equation}
where $k_1$ and $k_2$ are two randomly selected disjoint subsets of $K$ covariates $\{1,..., K\}$. Figure \ref{fig:fig_4} shows the histogram of event times for the Synthetic dataset. By applying an exponential function to the normally distributed features with additive Gaussian noise, the event times will be exponentially distributed with an average that depends on the parameters set $\boldsymbol{\beta}$ and $\boldsymbol{\alpha}$. Notice that the size of $\boldsymbol{\beta}$ and $\boldsymbol{\alpha}$ depend on the size of subsets $k_1$ and $k_2$ of the random variables, respectively. In this simulation, we considered $K=10$, $k_1=\{1,3,5,7\}$, and $k_2=\{2,4,6,8,9,10\}$ (this means the size of the parameter sets $\boldsymbol{\beta}$ and $\boldsymbol{\alpha}$ are 5 and 6, respectively). We generated 20000 data samples from this stochastic process. \\
In medical examinations and follow-ups of a patient, some of the clinical measurements may not have a significant contribution to the occurrence of an event of interest. Therefore, we call them insignificant clinical measurements. In the generated dataset, the time of the events is influenced by all of the clinical measurements defined in equation \ref{synthetic_data}. We added a few insignificant clinical measurements to the Synthetic dataset to test the feature selection capability of RSM. To add insignificant clinical measurements to the dataset, we considered a group of $M$ non-informative clinical measurements $\boldsymbol{x}_p=(x_p^{1},...,x_p^{M})$ that have no effect on the event times, $\boldsymbol{x}'=(x^{1},...,x^{K}, x_p^{0},...,x_p^{M})$. Here we considered five of such non-informative features for the dataset, $M=5$.
Real-life clinical datasets have some characteristics like containing censored and missing values. We considered such characteristics when generating our synthetic data to make the dataset more realistic \cite{lagakos1979general,leung1997censoring, ibrahim2012missing, sainani2015dealing, nazabal2020handling}. \\
\textbf{Right-censoring}: Right censoring is a common feature of medical datasets. Patients are frequently lost to the follow-ups. Consequently, their medical records are not gathered after the censoring time \cite{lagakos1979general,leung1997censoring}. To consider this real-world situation, we randomly selected half of the data, 10000 data samples, to be right-censored. Therefore, each data sample is represented by $(\boldsymbol{x}'_i, s_i, k_i)$, where $s_i$ indicates if the event time for a given data sample is right-censored ($s_i=1$) or not ($s_i=0$). $k_i$ shows the event time for the non-censored data samples and the lost to-follow-up time for the censored data samples.\\
\textbf{Missing values}: The other phenomenon that is frequently observed in medical data is the presence of missing values. In a longitudinal dataset, such as MIMIC-IV, only a few clinical measurements are recorded at a given time, leaving the rest of the clinical measurements unrecorded \cite{ibrahim2012missing, sainani2015dealing, nazabal2020handling}. It
has been noted that missing values and their missing patterns are often correlated with the target
labels, a.k.a., informative missingness, which leads to high missing rates for longitudinal datasets \cite{che2018recurrent}. We introduced such not-missing-at-random values to the Synthetic dataset by creating missing patterns for covariates that are to different extent correlated to labels. We introduced up to 45\% of such not-missing-at-random values to the dataset.  We also introduced up to 5\% missing-at-random values to clinical measurements. In sum, the overall missing rate of the Synthetic dataset is 50\%. \\
\subsection{The MIMIC-IV dataset}
MIMIC-IV is a large, freely-available database comprising de-identified health-related data associated with over 200,000 patients grouped into three modules: core, hosp, and ICU. The documentation of MIMIC-IV is available on its website \footnote{https://mimic.mit.edu/}. In this research, we use the ICU module that contains clinical measurements and outcome events of ICU patients, which can be used for survival analysis \cite{johnson2020mimic} (see appendix section for more details).
\section{Results}
\label{section:experiments}
In this section, we evaluate the ability of RSM in identifying the most relevant variables to the event time and ranking similar patients for MIMIC-IV and Synthetic datasets.
\subsection{Evaluation Approach/Study Design}
To evaluate the performance of a deep survival model, we considered the time-dependent Concordance-Index $\mathbb{C}^{td}(t)$ \cite{antolini2005time} and mean absolute error (MAE) \cite{chai2014root}. $\mathbb{C}^{td}(t)$ given by 
\begin{equation*}
    \mathbb{C}^{td}(t)=P(\hat{F}(t\mid x_i) > \hat{F}(t \mid x_j)\mid \delta_i =1, T_i<T_j, T_i\leq t),
\end{equation*}
where, $\hat{F}(t\mid x_i)$ is the estimated cumulative distribution function (CDF) of an event predicted by the model at time $t$, given clinical measurements $x_i$. $\delta_i$ is the event identifier and $\delta_i=1$ identifies uncensored data samples.
The time dependency in $\mathbb{C}^{td}(t)$ allows us to measure how effective the survival model is in capturing the possible changes in risk over time. We report $\mathbb{C}^{td}(t)$ at four  $25\%$, $50\%$, $75\%$, $100\%$ quantiles to roughly cover the whole event horizon.The MAE measure is given by $$MAE=\frac{\sum_{i=1}^{N} \mathbb{I}_{\{\delta_i ==1\}}(\|y_i - \hat{y}_i\|)}{\sum_{i=1}^{N} \mathbb{I}_{\{\delta_i ==1\}}}, $$ where $N$ is the sample size in each quantile, $\mathbb{I}_{\{\delta_i ==1\}}$ indicates the $i^{th}$ patient experienced an interested event, $y_i$ is the true event time, and $\hat{y_i}$ is the expected value of the predicted PDF. Note that the MAE measure is only calculated for uncensored patients.
\subsection{RSM results on the Synthetic dataset}
\begin{figure}[!tb]
		\centering
		\includegraphics[width=\linewidth]{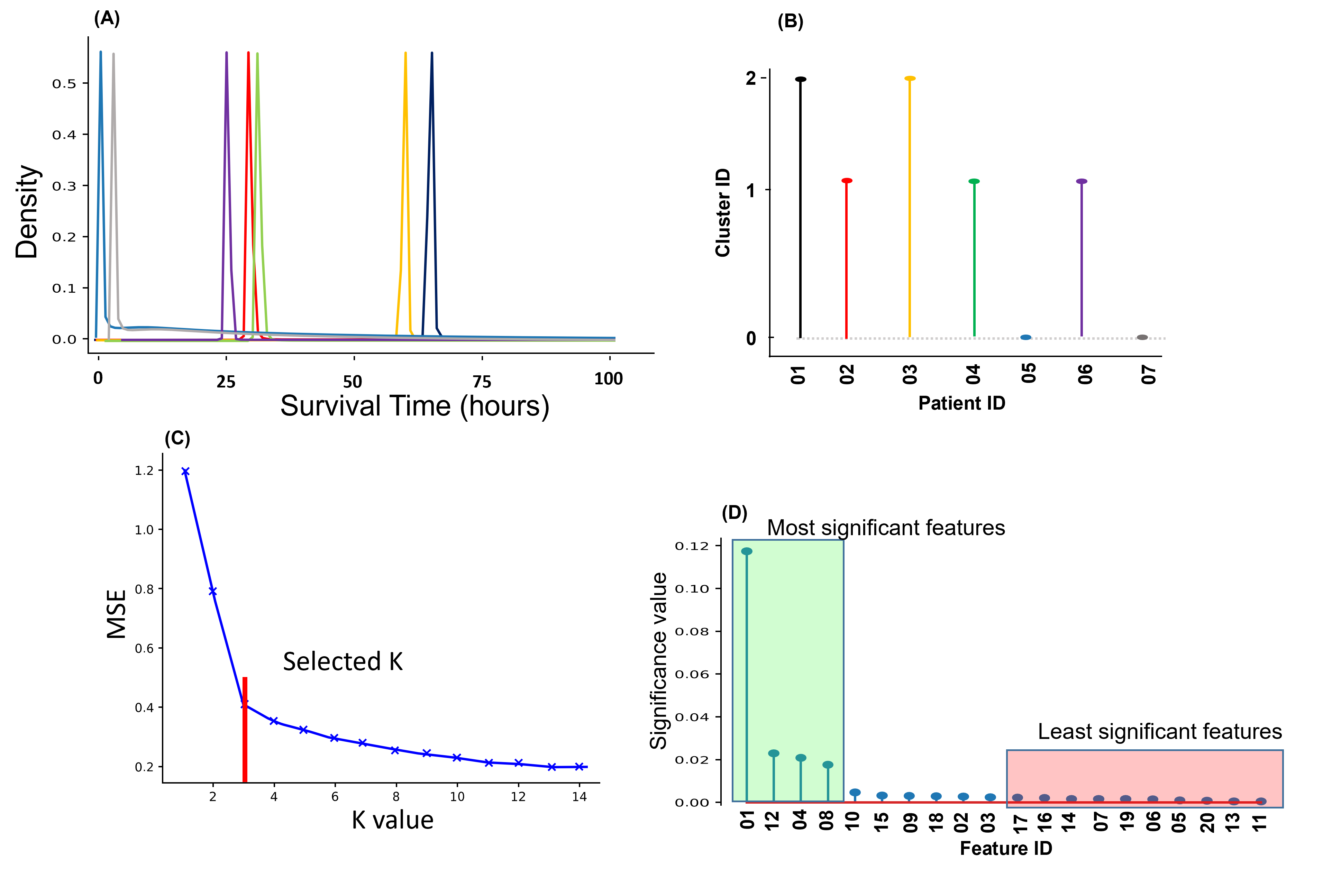}
		\caption{RSM identification results on the Synthetic dataset. A) Survival PDFs predicted by survival Seq2Seq for a group of 7 randomly selected patients, relabeled from 1 to 7. B) The clustering results for the group of 7 randomly selected patients whose PDFs are shown in (A).
		C) Finding the optimal number of clusters for K-Means. The selected value for K is 3 in this analysis. 
		D) Significant clinical measurements identified by RSM. The significant features are ranked based on their significant weights in a descending order. The most and least significant features identified using the KS statistical test are indicated by green and red boxes.}
		\label{fig:fig_Seq2Seq_result_synth}
	\end{figure}

To evaluate the performance of RSM, we applied it to the Survival Seq2Seq model described in \cite{survival_Seq2Seq}. Survival Seq2Seq has been recently proposed as a state-of-art model for survival analysis. The performance of Survival Seq2Seq on MIMIC-IV and Synthetic datasets is provided in Table \ref{tab:Seq2Seq_result}.
\begin{table}
\centering
\caption{The performance of Survival Seq2Seq \cite{survival_Seq2Seq} on Synthetic and MIMIC-IV datasets. Results are reported with 95\% confidence interval.}
\begin{tabular}{|l|l|llll|}
\hline
 & \begin{tabular}[c]{@{}l@{}}Performance\\ Measures\end{tabular} & \multicolumn{4}{c|}{Quantiles} \\ \hline
 &  & \multicolumn{1}{c|}{25\%} & \multicolumn{1}{c|}{50\%} & \multicolumn{1}{c|}{75\%} & \multicolumn{1}{c|}{100\%} \\ \hline
MIMIC\_IV & MAE & \multicolumn{1}{l|}{34.83±4.1} & \multicolumn{1}{l|}{37.06±4.6} & \multicolumn{1}{l|}{39.53±4.0} & 62.74±3.2 \\ \hline
MIMIC\_IV & CI & \multicolumn{1}{l|}{0.876±0.02} & \multicolumn{1}{l|}{0.882±0.02} & \multicolumn{1}{l|}{0.885±0.02} & 0.906±0.02 \\ \hline
Synthetic & MAE & \multicolumn{1}{l|}{11.85±0.6} & \multicolumn{1}{l|}{12.47±1.2} & \multicolumn{1}{l|}{14.01±1.4} & 15.54±1.8 \\ \hline
Synthetic & CI & \multicolumn{1}{l|}{0.874±0.00} & \multicolumn{1}{l|}{0.777±0.03} & \multicolumn{1}{l|}{0.772±0.05} & 0.807±0.08 \\ \hline
\end{tabular}

\label{tab:Seq2Seq_result}
\end{table}
As described in \cite{survival_Seq2Seq}, Survival Seq2Seq predicts a PDF for each data sample and event in the dataset. Figure  \ref{fig:fig_Seq2Seq_result_synth}.A shows an example of the outcome of Survival Seq2Seq for a group of 7 simulated patients.
Figure \ref{fig:fig_Seq2Seq_result_synth}.B shows the outcome of clustering on survival PDFs estimated by survival Seq2Seq. The optimal number of the clusters is identified by measuring the mean of squared distances, as shown in Figure \ref{fig:fig_Seq2Seq_result_synth}.C. As previously mentioned, we intentionally added non-informative clinical measurements to the Synthetic dataset. Figure \ref{fig:fig_Seq2Seq_result_synth}.D shows RSM can successfully identify those non-informative clinical measurements and exclude them from the rest of the features by assigning smaller weights to them. We had considered features with IDs 16 to 20 as non-informative, where RSM successfully identifies 4 of them as the least significant features. In addition, none of the non-informative features are identified by RSM as the most significant features. This shows that RSM can correctly identify the significant features of the Synthetic dataset.
\subsection{RSM results on the MIMIC-IV dataset}
\begin{figure}[!tb]
		\centering
		\includegraphics[width=\linewidth]{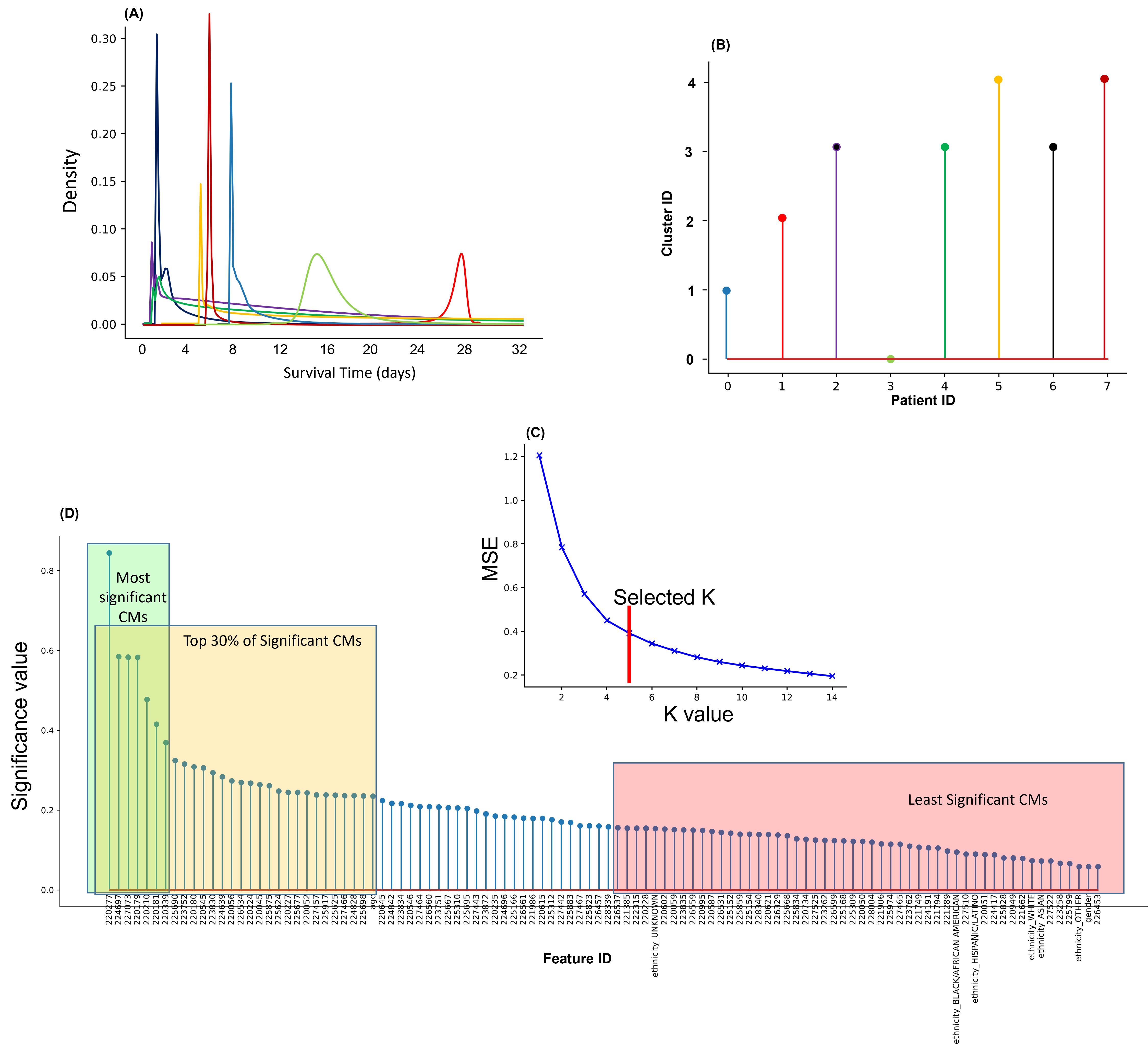}
		\caption{The RSM results for the MIMIC-IV dataset. A) The survival PDFs predicted by survival Seq2Seq for a group of 10 random patients. Patients are labelled from 1 to 10. B) Clustering results for a group of 10 randomly selected patients whose PDFs were shown in (A) with the same colors.
		C) Finding the optimal number of the clusters for K-Means. The selected value for K is 5 in this analysis. 
		D) The most significant and top 30\% of the significant clinical measurements identified by RSM are shown and grouped by different boxes.  Due to the sheer number of clinical measurements in MIMIC-IV, we only show the significance values of the most significant features. The significant features are ranked based on their significant weights in a descending order.}
		\label{fig:fig_Seq2Seq_result_MIMIC}
	\end{figure}

\begin{figure}[!tb]
		\centering
		\includegraphics[width=\linewidth]{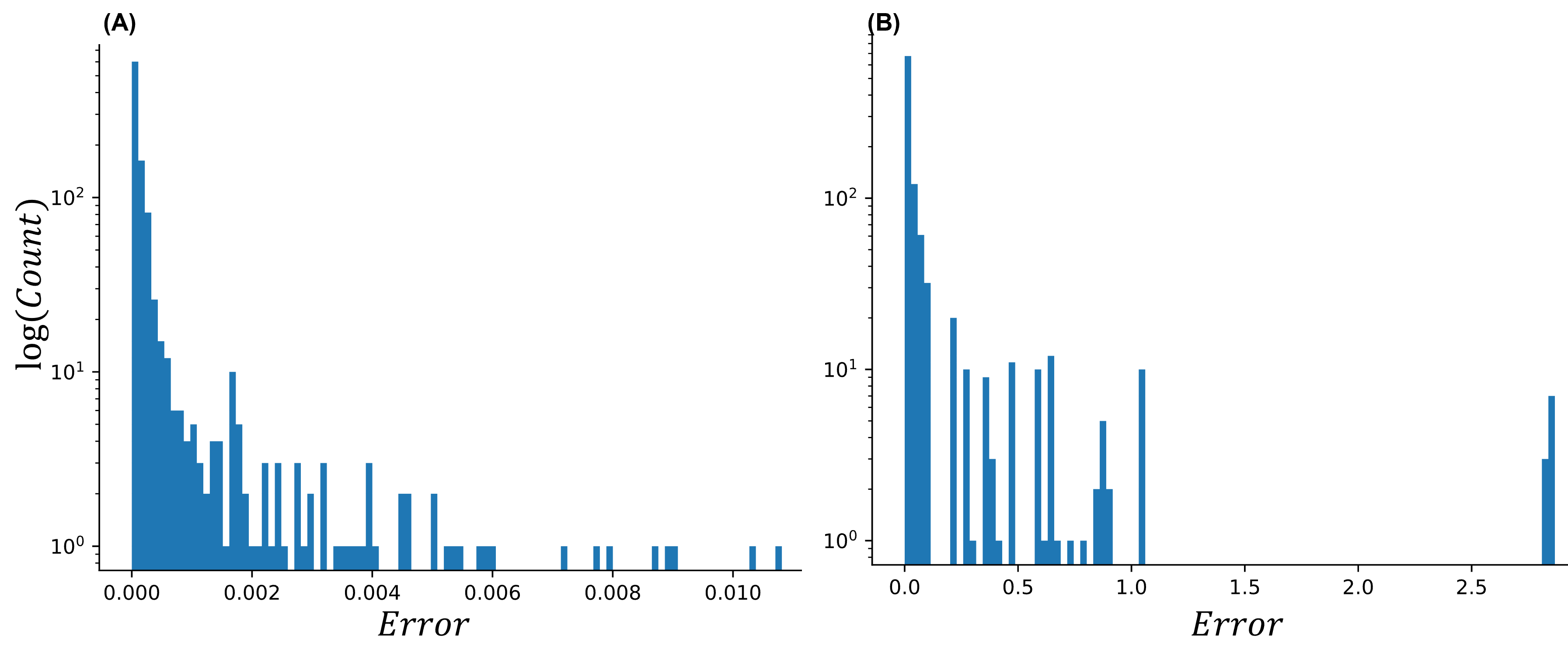}
		\caption{The histogram of similarity score error for the test patients with A) kernel-PCA and B) PCA analysis.}
		\label{fig:fig06}
	\end{figure}

\begin{figure}[!tb]
		\centering
		\includegraphics[width=\linewidth]{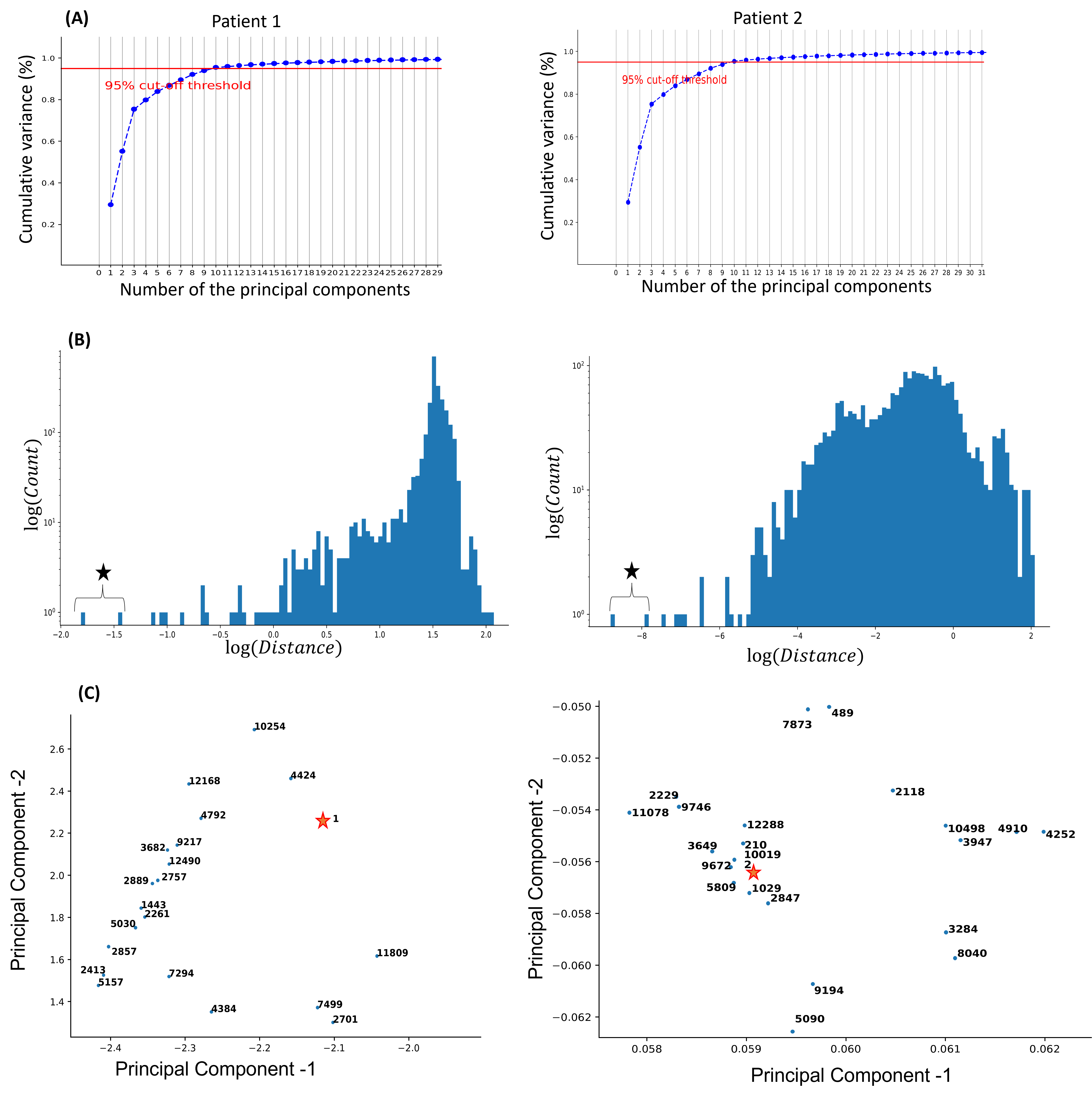}
		\caption{RSM identifies similar patients based on their survival PDFs and most significant clinical measurements.
		A) PCA analysis for 95\% cut-off threshold of the cumulative variance identification for the patients with IDs 1 (left plot) and 2 (right plot).
		B) Histogram of the logarithm of the score for the most representative principal components for the patients and associated similar patients. The score range of the most similar patients are identified by a star.
		C) Visualization of the first two principal components for the most similar patients those selected from the histogram of distances (B)) for patients with IDs 1 (left plot) and 2 (right plot).}
		\label{fig:fig07}
	\end{figure}

The performance of RSM is evaluated using MIMIC-IV dataset as shown in Figure \ref{fig:fig_Seq2Seq_result_MIMIC}.
Table \ref{tab:sign_perf_MIMIC} shows the top 30\% of the most significant clinical measurements of MIMIC-IV. Medical references that confirm the significance of the clinical measurements identified by RSM are also provided in that table.
\begin{table}[tb]
\small
\centering
\caption{Top 30\% of the most significant clinical measurements of MIMIC-IV in survival analysis, identified by RSM.}
\begin{tabular}{|l|ll|}
\hline
\textit{\begin{tabular}[c]{@{}l@{}}Significance\\  percent\end{tabular}} & \multicolumn{2}{c|}{Clinical Measurements} \\ \hline
1\% & \multicolumn{1}{l|}{\begin{tabular}[c]{@{}l@{}} 1-O2 saturation \\ pulseoxymetry \cite{vold2015low}\end{tabular}} & \multicolumn{1}{l|}{\begin{tabular}[c]{@{}l@{}} 2-Mean Airway \\ Pressure \cite{sahetya2020mean}\end{tabular}}  \\ \hline
2-3\% & \multicolumn{1}{l|}{3-Anion gap \cite{zhang2022value}} & \begin{tabular}[c]{@{}l@{}} 4-Non Invasive Blood\\  Pressure systolic \cite{lacson2019use}\end{tabular} \\ \hline
4-5\% & \multicolumn{1}{l|}{5-Respiratory Rate \cite{kang2020machine}} & \begin{tabular}[c]{@{}l@{}}6-Non Invasive Blood\\  Pressure mean \end{tabular} \\ \hline
6-7\% & \multicolumn{1}{l|}{7-PEEP set} & 8-Total Bilirubin \cite{chen2021association} \\ \hline
8-9\% & \multicolumn{1}{l|}{\begin{tabular}[c]{@{}l@{}}9-Non-Invasive Blood \\ Pressure Alarm - Low\end{tabular}} & \begin{tabular}[c]{@{}l@{}}10-Non Invasive Blood\\  Pressure diastolic \cite{greenberg2006blood}\end{tabular} \\ \hline
10-11\% & \multicolumn{1}{l|}{11-Hematocrit (serum) \cite{erikssen1993haematocrit}} & 12-PH (Arterial) \cite{kang2020machine} \\ \hline
12-13\% & \multicolumn{1}{l|}{13-Daily Weight \cite{kang2020machine}} & \begin{tabular}[c]{@{}l@{}}14-Arterial Blood\\  Pressure Alarm - Low\end{tabular} \\ \hline
14-15\% & \multicolumn{1}{l|}{Sodium (whole blood)} & Arterial O2 pressure \\ \hline
16-17\% & \multicolumn{1}{l|}{15-Heart Rate \cite{chen2022development}} & 16-Gentamicin \cite{chen2022development} \\ \hline
18-19\% & \multicolumn{1}{l|}{17-BUN \cite{beier2011elevation}} & 18-Arterial O2 Saturation \\ \hline
20-21\% & \multicolumn{1}{l|}{19-Phosphorous \cite{kestenbaum2005serum}} & \begin{tabular}[c]{@{}l@{}}20-Arterial Blood\\  Pressure mean\end{tabular} \\ \hline
22-23\% & \multicolumn{1}{l|}{21-Platelet Count \cite{msaouel2014abnormal}} & 22-TPN without Lipids \\ \hline
24-25\% & \multicolumn{1}{l|}{23-Calcium non-ionized\cite{miller2010association}} & 24-PTT \cite{reddy1999partial} \\ \hline
26-27\% & \multicolumn{1}{l|}{25-Arterial Base Excess \cite{hamed2019base}} & 26-TCO2 (calc) Arterial \cite{wayne1995use} \\ \hline
28-30\% & \multicolumn{1}{l|}{27-age \cite{ferreira2022two}} & 28-Sodium (serum) \cite{vaa2011influence} \\ \hline
\end{tabular}
\label{tab:sign_perf_MIMIC}
\end{table}
We also trained the survival Seq2Seq model using only the top 30\% of the most significant clinical measurements identified by RSM, where the outcome is presented in Table \ref{tab:Seq2Seq_result_MIMIC_30}. Our objective was to investigate if training Survival Seq2Seq using the most significant features would result in an outcome close to the original MIMIC-IV outcome reported in Table \ref{tab:Seq2Seq_result}. In other words, we wanted to verify if the significant features that RSM identifies indeed carry the information that is most relevant to survival analysis. It can be observed from Table \ref{tab:Seq2Seq_result_MIMIC_30} that the performance of Survival Seq2Seq drops slightly compared to the original results reported in Table  \ref{tab:Seq2Seq_result}. This verifies that RSM is able to accurately identify the most significant clinical measurements for MIMIC-IV.
\begin{table}[tb]
\caption{Performance of Survival Seq2Seq \cite{survival_Seq2Seq} trained on the top 30\% of the most significant clinical measurements of MIMIC-IV. Results are reported with 95\% confidence interval.}
\begin{tabular}{|l|llll|}
\hline
\begin{tabular}[c]{@{}l@{}}Performance\\ Measures\end{tabular} & \multicolumn{4}{c|}{Quantiles} \\ \hline
 & \multicolumn{1}{c|}{25\%} & \multicolumn{1}{c|}{50\%} & \multicolumn{1}{c|}{75\%} & \multicolumn{1}{c|}{100\%} \\ \hline
MAE & \multicolumn{1}{l|}{34.78±7.41} & \multicolumn{1}{l|}{36.12±7.35} & \multicolumn{1}{l|}{38.58±6.04} & 64.0±5.72 \\ \hline
CI & \multicolumn{1}{l|}{0.847±0.035} & \multicolumn{1}{l|}{0.846±0.013} & \multicolumn{1}{l|}{0.840±0.019} & 0.651±0.013 \\ \hline
\end{tabular}

\label{tab:Seq2Seq_result_MIMIC_30}
\end{table}
For the MIMIC-IV dataset with hundreds of numerical and categorical clinical measurements, as described in section \ref{PCA_desc_section}, we suggest using PCA or kernel-PCA methods to measure the similarity score between patients. Figure \ref{fig:fig06} shows the histogram of error for the similarity score between the test patients. As expected, kernel-PCA shows smaller errors due to the presence of nonlinear relationships between patients' clinical measurements. 
In the end, RSM ranks the most similar patients to a patient of interest based on the score measure identified by the kernel-PCA. As an example, the most similar patients to patients with IDs 1 and 2 are shown in Figure \ref{fig:fig07}.
\section{Discussion}
\label{section:Discussion}
In this study, we proposed RSM, a framework that identifies patients with similar clinical measurements and outcomes to a patient of interest, and therefore verifies the predictions of deep survival models. We applied RSM to the predicted survival PDFs of the Survival Seq2Seq model trained on a synthetic dataset to validate the ability of RSM in recognizing significant clinical measurements and identifying the data samples that are most similar to a given data sample. After validating these capabilities, we tested RSM on Survival Seq2Seq trained on MIMIC-IV to justify the predictions of Survival Seq2Seq. To the best of our knowledge, this is the first time a framework has been specifically designed to interpret the outputs of a trained deep survival model.\\
\subsubsection*{Limitations} Despite adding the interpretation capability to deep survival models, a few limitations can affect the performance of RSM. RSM uses a deep model for performing survival analysis. As a result, the precision of the deep survival model has a direct impact on the performance of RSM. RSM cannot properly interpret the outcome of a deep survival that model suffers from poor predictions. The other limitation of RSM is that its overall performance is bounded by the individual performance of its several interpretation units such as the similarity measure, clustering model, $L_1$ significant feature selection, and statistical tests. One can replace each of the mentioned components of RSM with a more advanced algorithm to achieve a higher overall performance for RSM. For example, to achieve a better performance in clustering, a clustering algorithm based on deep unsupervised learning \cite{dilokthanakul2016deep} can be used. Upgrading a single or all units employed in RSM is a subject for our future research. In addition, identification of significant feature in RSM is based on the training cohort. As health care industry pursues individualized services, a case-specific significant feature identification is potentially more desirable.

\begin{appendices}
\section{MIMIC-IV database}
The MIMIC-IV database contains health related data of ICU patients of Beth Israel Deaconess Medical Center, between 2008 and 2019. There are a total of number of 71791 distinct ICU admission records with an average ICU stay of 4 days. This dataset includes vital sign measurements, laboratory test results, medications, imaging report of the patients, stored in separate tables. For mortality prediction, relevant covariates has been extracted from the following three tables: 1) INPUTEVENTS (continuous infusions or intermittent administrations), 2) OUTPUTEVENTS (patient outputs including urine, drainage, and so on), and 3) CHARTEVENTS (Patient's routine vital signs and any additional information relevant to their care during ICU stay). We selected a total of 108 covariates from these three tables. These covariates were selected based on the feedback from our medical team, as well as applying conventional feature selection techniques on the dataset. The number of patients after feature selection dropped to 66363 with an uncensored (deceased patients) rate of about $12\%$. The following table lists the total number of covariates and selected number of covariates from each table.
\begin{table}
    \caption{\small \label{tbl:MIMIC-TABLE-DIST} \small MIMIC-IV tables with their corresponding number of total and selected covariates.}
    \begin{tabular}{|c|c|c|}
        \hline
        \textbf{Table} & \textbf{Total \# of Covariates}  & \textbf{\# of Selected Covariates} \\
        \hline
        INPUTEVENTS & 282 & 30 \\
        \hline
        OUTPUTEVENTS & 69 & 5 \\
        \hline
        CHARTEVENT & 1566 & 73 \\
        \hline
        \textbf{Total} & 1917 & 108 \\
        \hline
    \end{tabular}
\end{table}
\subsection*{Data Pre-Processing Considerations}
\begin{itemize}
\item Patients with the following diagnosis are excluded from the data: Sudden Infant Death synd, unattended death, maternal death affecting fetus or newborn, fetal death from asphyxia or anoxia during labor, intrauterine death.
\item Invalid measurements were removed from the data using the provided WARNING and ERROR columns. 
\item Survival time was defined as the period between the time of admission and time of death (for patients who died in hospital), or time of discharge (for censored patients).
\item Time of observation is defined as the time of recording of the measurement with admission time as the baseline.
\end{itemize}
\end{appendices}
\subsubsection*{Acknowledgment}
We would like to sincerely thank Health Canada for their kind support for funding the challenge "Machine learning to improve organ donation rates and make better matches" (Challenge ID: 201906-F0022-C00008). This challenge aims to improve the quality of organ matchmaking and increase the pool of Donation after Circulatory Death (DCD) donors.
\section*{Declarations}
\subsubsection*{Conflict of Interests} The author declares that he has no known competing financial interests or personal relationships that could have appeared to influence the work reported in this paper.
\bibliography{sn-bibliography}
\end{document}